\documentclass[11pt,a4paper]{article}
\usepackage[hyperref]{eacl2021}
\usepackage{times}
\usepackage{latexsym}

\usepackage{microtype}

\usepackage{graphicx}
\usepackage{amsfonts, amsmath, amsthm, amssymb}
\usepackage{xspace}
\usepackage{float}
\usepackage{multirow}
\usepackage[T1]{fontenc}
\usepackage{enumitem}
\usepackage{booktabs}
\usepackage{comment}
\usepackage[export]{adjustbox}
\usepackage{array}
\usepackage{xcolor}
\usepackage{soul}
\usepackage{caption}
\usepackage{subcaption}

\definecolor{myblue}{HTML}{1E88E5}
\definecolor{mygreen}{HTML}{19A717}
\definecolor{mypink}{HTML}{D88AC2}
\definecolor{mydarkgreen}{HTML}{004D40}
\definecolor{myorange}{HTML}{FFC107}
\definecolor{mypurple}{HTML}{D81B60}

\usepackage{cleveref}
\crefname{section}{\S}{\S\S}
\crefname{table}{Table}{}
\crefname{figure}{Figure}{}
\crefname{algorithm}{Algorithm}{}
\crefname{equation}{eq.}{}
\crefname{appendix}{App.}{}
\crefname{prop}{Proposition}{}
\crefname{thm}{Theorem}{}
\crefformat{section}{\S#2#1#3}  

\newcommand{\butd}{\textsc{butd}\xspace}
\newcommand{\butr}{\textsc{butr}\xspace}

\newcommand{\trm}{$\mathcal{M}^2\textsc{-TRM}$\xspace}
\newcommand{\butdb}{\textbf{\textsc{BUTD}}\xspace}
\newcommand{\butrb}{\textbf{\textsc{BUTR}}\xspace}
\newcommand{\trmb}{$\boldsymbol{\mathcal{M}^2}\textbf{\textsc{-TRM}}$\xspace}
\newcommand{\bleu}{\textsc{BLEU}\xspace}
\newcommand{\meteor}{\textsc{METEOR}\xspace}
\newcommand{\cider}{\textsc{CIDEr}\xspace}
\newcommand{\spice}{\textsc{SPICE}\xspace}
\newcommand{\ciderd}{\textsc{CIDEr-D}\xspace}
\newcommand{\bertscore}{\textsc{BERTScore}\xspace}
\newcommand{\rfive}{$\mathrm{R@5}$\xspace}
\newcommand{\recallk}{$\mathrm{Recall@K}$\xspace}
\newcommand{\recall}{\textsc{Recall}\xspace}
\newcommand{\idle}{\textsc{idle}\xspace}
\newcommand{\idlex}{\textsc{<idle>}\xspace}
\newcommand{\chunk}{\textsc{chunk}\xspace}
\newcommand{\pos}{\textsc{pos}\xspace}
\newcommand{\dep}{\textsc{dep}\xspace}
\newcommand{\ccg}{\textsc{ccg}\xspace}

\newcommand{\plan}{\textsc{sequential}\xspace}
\newcommand{\inter}{\textsc{interleave}\xspace}
\newcommand{\multi}{\textsc{multi-task}\xspace}
\newcommand{\butrmean}{$\textsc{butr}_{mean}$\xspace}

\newcommand{\butrwt}{$\textsc{butr}_{weight}$\xspace}

\usepackage{todonotes}
\makeatletter
\newcommand*\iftodonotes{\if@todonotes@disabled\expandafter\@secondoftwo\else\expandafter\@firstoftwo\fi}  
\makeatother

\usepackage{euflag}

\aclfinalcopy 

\title{The Role of Syntactic Planning in Compositional Image Captioning}

\author{Emanuele Bugliarello \and Desmond Elliott\\
  Department of Computer Science\\
  University of Copenhagen \\
  \texttt{\{emanuele,de\}@di.ku.dk}}

\date{}

\begin{document}
\maketitle

\begin{abstract}
Image captioning has focused on generalizing to images drawn from the same distribution as the training set, and not to the more challenging problem of generalizing to different distributions of images. 
Recently, \newcite{nikolaus-etal-2019-compositional} introduced a dataset to assess compositional generalization in image captioning, where models are evaluated on their ability to describe images with unseen adjective--noun and noun--verb compositions. 
In this work, we investigate different methods to improve compositional generalization by planning the syntactic structure of a caption.
Our experiments show that jointly modeling tokens and syntactic tags enhances generalization in both RNN- and Transformer-based models, while also improving performance on standard metrics.
\end{abstract}

\section{Introduction}

Image captioning is a core task in multimodal NLP, where the aim is to automatically describe the content of an image in natural language.
To succeed in this task, a model first needs to recognize and understand the properties of the image. 
Then, it needs to generate well-formed sentences, requiring both a syntactic and a semantic knowledge of the language \cite{hossain2019comprehensive}. 
Deep learning techniques are the standard approach to tackling this problem: images are represented by visual features extracted from Convolutional Neural Networks (\textit{e.g.} \citealt{he2016deep}), and sentences are generated by conditioning Recurrent Neural Networks (\textit{e.g.} \citealt{hochreiter1997long}), or Transformers \cite{vaswani2017attention} on the extracted visual features.

While deep neural networks achieve impressive performance in a variety of applications, including image captioning, their ability to demonstrate compositionality, defined as the algebraic potential to understand and produce novel combinations from known components \cite{loula-etal-2018-rearranging}, has been questioned.
Semantic compositionality of language in neural networks has attracted interest in the community~\cite{NIPS2014_5551,pmlr-v80-lake18a,baroni2019linguistic} as compositionality is conjectured to be a core feature not only of language but also of human thought \cite{fodor2002compositionality}.

In image captioning, improving compositional generalization is a fundamental step towards generalizable systems that can be employed in daily life.
To this end, \newcite{nikolaus-etal-2019-compositional} recently introduced a compositional generalization dataset where models need to describe images that depict unseen compositions of primitive concepts. 
For example, models are trained to describe images with ``white'' entities and all types of ``dog'' concepts but never the adjective--noun composition of ``white dog.'' 
In their dataset, models are evaluated on their ability to caption images depicting the unseen composition of held out concepts. 
Their study suggests that RNN-based captioning models do not compositionally generalize, and that this is primarily attributable to the language generation component.

In this paper, we study the potential for syntax to improve compositional generalization in image captioning by combining syntactic planning and language generation in a single model.
Our study is inspired by the traditional Natural Language Generation (NLG) framework \cite{reiter1997building}, where NLG is split into three distinct steps: text planning, sentence planning, and linguistic realization.
While state-of-the-art captioning models typically proceed directly from visual features to sentence generation, we hypothesize that a model that plans the structure of a sentence as an intermediate step will improve compositional generalization. 
A model with a planning step can learn the high-level structure of sentences, making it less prone to overfitting the training data.

Specifically, we explore three methods for integrating syntactic planning into captioning in our experiments: (a) pre-generation of syntactic tags from the image, (b) interleaved generation of syntactic tags and words \cite{nadejde-etal-2017-predicting}, and (c) multi-task learning with a shared encoder that predicts syntactic tags or words \cite{currey-heafield-2019-incorporating}. 
We do so while also empirically investigating four different levels of syntactic granularity. 

The main findings of our experiments are that:
\begin{itemize}[topsep=1pt]
    \setlength\itemsep{1pt}
    \item jointly modeling syntactic tags and tokens leads to improvements in Transformer-based \cite{cornia2020m2} and RNN-based \cite{anderson2018} image captioning models;
    \item although the effectiveness of each syntactic tag set varies across our explored approaches, the widely-used \emph{chunking} tag set never outperforms syntactic tags with finer granularity;
    \item compositional generalization is affected by directly mapping from image representation to tokens because performance can be improved by interleaving a dummy tag with no meaning;
    \item interleaving syntactic tags with tokens leads to a loss in performance for retrieval systems.
\end{itemize}
Finally, we also propose an attention-driven image--sentence ranking model, which makes it possible to adaptively combine syntax within the re-scoring approach of \citet{nikolaus-etal-2019-compositional} to further improve compositional generalization in image captioning.
\section{Planning Image Captions}

\begin{figure*}[t]
  \centering
  \includegraphics[width=0.9\linewidth, trim={0cm 3.5cm 0.cm 3.6cm}, clip]{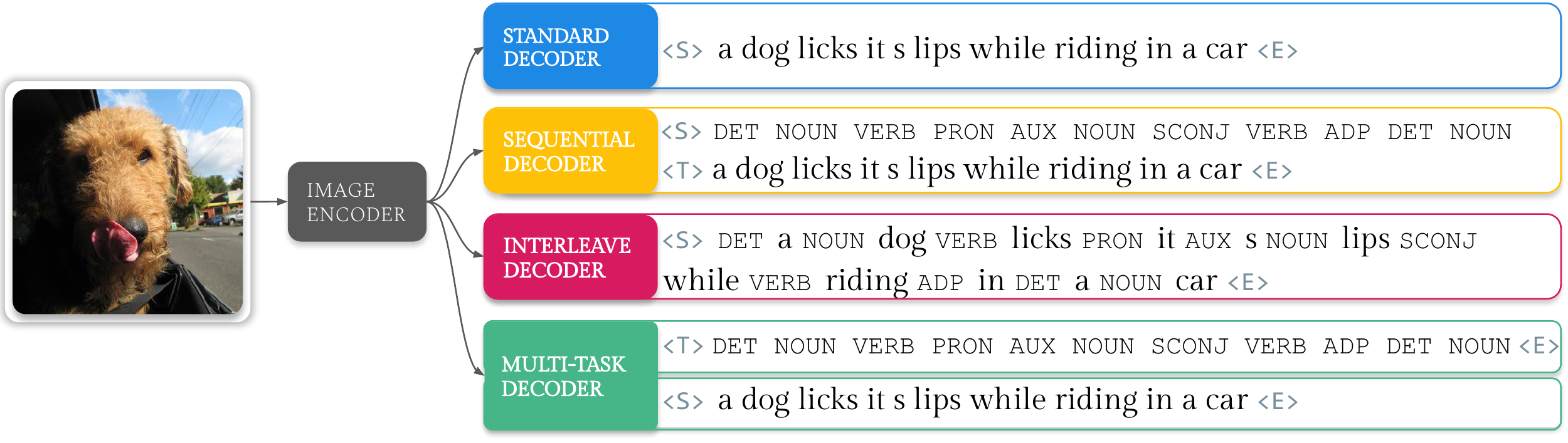}
  \caption{Approaches to syntactically plan image captioning (with \pos tags). \textsc{Standard} captioning systems directly generate a sequence of surface forms (i.e. words). \plan generates a sequence of syntactic tags, followed by a sequence of surface forms. \inter alternates syntactic tags and surface forms. \multi generates either a sequence of syntactic tags or a sequence of surface forms from a shared decoder.}
  \label{fig:syncap}
\end{figure*}

Natural language generation has traditionally been framed in terms of six basic sub-tasks: content determination, discourse planning, sentence aggregation, lexicalization, referring expression generation and linguistic realization~\cite{reiter1997building}.
Within this framework, a three-stage pipeline has emerged~\cite{reiter-1994-consensus}:
\begin{itemize}[noitemsep,topsep=1pt]
    \item \textbf{Text Planning:} combining content determination and discourse planning.
    \item \textbf{Sentence Planning:} combining sentence aggregation, lexicalization and referring expression generation to determine the structure of the selected input to be included in the output.
    \item \textbf{Linguistic Realization:} this stage involves syntactic, morphological and orthographic processing to produce the final sentence.
\end{itemize}

Early methods for image captioning drew inspiration from this framework; for example, the \textsc{Midge} system \citep{mitchell-etal-2012-midge-generating} features explicit steps for content determination, given detected objects, and sentence aggregation based on local and full phrase-structure tree construction, and \textsc{Treetalk} composes tree fragments using integer linear programming \citep{kuznetsova2014}. 
More recently, \newcite{wang2017skeleton} propose a two-stage algorithm where the skeleton sentence of the caption (main objects and their relationships) is first generated, and then the attributes for each object are generated if they are worth mentioning.
In contrast, the majority of neural network models are based on the encoder-decoder framework \citep{sutskever2014} of learning a direct mapping from different granularities of visual representations \citep{vinyals2015show,pmlr-v37-xuc15,anderson2018} to language model decoders based on RNNs \citep{vinyals2015show} or Transformers \cite{guo2020normalized,cornia2020m2}.

\subsection{Motivation}
In this paper, we explore whether image captioning models can be improved by explicitly modeling sentence planning as an intermediate step between content determination and linguistic realization. 
In particular, we study the use of syntactic tags in enriching the sentence planning step to improve compositional generalization.
In the compositional image captioning task, models are tasked with describing images that depict unseen combinations of adjective--noun and noun--verb constructions (see \citealt{nikolaus-etal-2019-compositional} for a more detailed description of this task). \citet{nikolaus-etal-2019-compositional} presented a model that improves generalization with a jointly trained discriminative re-ranker, whereas here, we investigate the role of sentence planning via syntax.

From a psycholinguistic perspective \citep{griffin2000eyes,coco2012scan}, there is evidence that humans make plans about how to describe the visual world: they first decide what to talk about (analogous to content determination), then they decide what they will say (a sentence planning phase), and finally, they produce an utterance (linguistic realization). 
We hypothesize that, analogously to humans, neural network decoders will also find it useful to make such sentence plans.

From a machine learning perspective, the use of syntactic structure can mitigate the bias introduced by the maximum likelihood training of neural network image captioning models. 
Recall that in the context of image captioning, the optimization objective consists of maximizing the likelihood:
\begin{equation}
    \mathcal{L} = \prod_{t=1}^T \mathbb{P}(y_t | y_{<t}, v),
\end{equation}
where $v$ denotes the visual features (either a single vector or a set of vectors extracted from an image).

In a standard generation task, a model learns to predict the next token based on what it has observed so far.
This is especially limiting when it is evaluated on unseen combinations of adjective--noun and noun--verb constructions in the compositional generalization task (i.e. data points that fall outside the training distribution).
In fact, models are not explicitly asked to learn \emph{word classes} nor how to connect them to form novel combinations.
Whereas, if a system also models syntax, it can assign higher probability to ``white dog'' if it expects to generate a sequence with an adjective followed by a noun. 

\subsection{Planning Approaches}

We investigate three approaches to jointly modeling tokens and syntactic tags: syntax-driven sequential caption planning (\plan), syntax-interleaved caption generation (\inter), and syntax and caption generation via multi-task learning (\multi). See \cref{fig:syncap} for an overview. 

\paragraph{\textbf{\plan:}}
    Our first approach closely follows the traditional NLG pipeline and it is related to the \textit{text planning} stage defined above, although limited to sentence-level rather than to a full discourse.
    Here, a model plans, through syntactic tags, the order of the information to be presented.
    Specifically, the model is required to generate a sequence whose first $T$ outputs represent the underlying syntactic structure of the sentence before subsequently generating the corresponding $T$ surface forms.

\paragraph{\textbf{\inter:}} 
    Our second approach consists of interleaving syntactic tags and tokens during generation, which means a syntactic tag and its realization are next to each other, removing the pressure for a model to successfully track long-range dependencies between tags and tokens.
    Moreover, this allows for a more flexible planning, where the model can adapt the sentence structure based on the previously generated tags and tokens.
    In particular, a model can break bi-gram dependencies and learn narrower distributions over the next word based on the current syntactic tag.
    For instance, if we consider part-of-speech tags, the model learns that only a subset of the vocabulary corresponds to nouns, and another subset to adjectives, and so on.

\paragraph{\textbf{\multi:}} 
    Our last approach is based on multi-task learning, where a model produces either a sequence of tokens (main task) or syntactic tags (secondary task).
    We draw on the simple and effective approach of~\newcite{currey-heafield-2019-incorporating}, proposed for neural machine translation (NMT).
    In the NMT framework, the source sentence was prepended a task-specific tag, which led the decoder to either predict the translation of the source sentence or the syntactic tags of the source sentence.
    We adapt this to image captioning by setting the first token to either a start-of-syntax token (\texttt{<T>}) or  start-of-sentence token (\texttt{<S>}) and then generating tags or tokens, respectively.
    Compared to the other approaches, \multi allows the model to learn both types of forms at the same position.
    While this approach does not double sequence length, it doubles the number of sequences per training epoch.

\subsection{Syntactic Granularity}
In addition to the three approaches of realizing sentence planning, we investigate the effects of different syntactic tags from a coarse to fine granularity. We experiment with the following tags:
\begin{itemize}[noitemsep,topsep=1pt]
    \item \textbf{\chunk:} Also known as \textit{shallow parsing}, chunks are syntactic tags that model phrasal structure in a sentence, such as noun phrases (NP) and verb phrases (VP).
    \item \textbf{\pos:} Part-of-speech tags are specific lexical categories to which words are assigned, based on their syntactic context and role, such as nouns (N) and adjectives (ADJ).
    \item \textbf{\dep:} Dependency-based grammars model the structure as well as the semantic dependencies and relationships between words in a sentence.
    In this study, we consider the dependency labels assigned to each word, such as adjectival modifiers (amod), which denote any adjective that modifies the meaning of a noun.
    \item \textbf{\ccg:} Combinatory categorial grammar \citep{STEEDMAN2006610} is based on combinatory logic and provides a transparent interface between surface syntax and the underlying semantic representation.
    For example, the syntactic category assigned to ``sees'' is ``(S\textbackslash NP)/NP'', denoting it as a transitive verb that will be followed by a noun phrase.
\end{itemize}
We also study the merit of breaking bi-gram dependencies for the \inter approach by tagging each word with a synthetic tag \idlex.
We hypothesize this approach would not give any benefits in any metric, as attention-based models can simply learn to ignore these pseudo-tags.
\section{Experimental Setup}

\paragraph{Data}

We use training and evaluation sets such that paradigmatic gaps exist in the training set.
That is, for a concept pair $\{c_i, c_j\}$, the validation $\mathcal{D}_{val}$ and test $\mathcal{D}_{test}$ sets only contain images in which at least one of the captions contains the pair of concepts, while the complementary set -- where concepts $c_i$ and $c_j$ can only be observed independently -- is used for training $\mathcal{D}_{train}$.
Following \citet{nikolaus-etal-2019-compositional}, we select the same $24$ adjective--noun and verb--noun concept pairs, and split the English COCO dataset~\cite{lin2014microsoft} into four sets, each containing six held out concept pairs.

\paragraph{Pre-processing}
We first lower-case and strip away punctuation from the captions.
We then use StanfordNLP~\cite{qi2018universal} to tokenize and lemmatize the captions, and to extract universal \pos tags and syntactic dependency relations.
For IOB-based chunking, we train a classifier-based tagger on \textit{CoNLL2000} data~\cite{tjong-kim-sang-buchholz-2000-introduction} using NLTK~\citep{bird2009natural}.
Finally, we use the A* \ccg parsing model by \newcite{yoshikawa-etal-2017-ccg} with ELMo embeddings \cite{peters-etal-2018-deep} to extract \ccg tags.
Visual features are extracted from $36$ regions of interest in each image using Bottom-Up attention \cite{anderson2018} trained on Visual Genome~\citep{krishna2017visual}.

\paragraph{Evaluation}
Following~\citet{nikolaus-etal-2019-compositional}, we evaluate compositional generalization with \recallk.
Given $K$ generated captions for each of the $M$ images in an evaluation set, $\{\langle s_1^1, \dots, s_K^1 \rangle, \dots, \langle s_1^M, \dots, s_K^M \rangle \}$, the recall of the concept pairs is given by:
\begin{equation}
    \mathrm{Recall@K} = \frac{|\{\langle s_k^m \rangle | \exists k : s_k^m \in \mathcal{C}\}|}{M},
\end{equation}
where $s_k^m$ denotes the $k$-th generated caption for image $m$ and $\mathcal{C}$ is the set of captions which contain the expected concept pair and in which the adjective or the verb is a dependent of the noun.

In addition, we use \texttt{pycocoeval} to score models on the common image captioning metrics: \meteor (M; \citealt{denkowski-lavie-2014-meteor}), \spice (S; \citealt{anderson2016}), \cider (C; \citealt{Vedantam2015}), and \bleu (B; \citealt{papineni-etal-2002-bleu}); and the recent multi-reference \bertscore (BS; \citealt{yi-etal-2020-improving}).
In particular, we report the average recall across all concept pairs, the average across the four splits for each score in \texttt{pycocoeval}, and the average across all captions for \bertscore.

\paragraph{Models} We evaluate three models:
\begin{itemize}[noitemsep,topsep=1pt]
    \item \butdb: Bottom-Up and Top-Down attention \cite{anderson2018}, a strong and widely-employed RNN-based captioning system.
    \item \butrb: Bottom-Up and Top-Down attention with Ranking \cite{nikolaus-etal-2019-compositional}, an RNN-based, multi-task model trained for image captioning and image--sentence ranking that achieves state-of-the-art performance in the compositional generalization task.\footnote{We denote as BUTR the model that uses the re-ranking module (BUTR+RR in \citealt{nikolaus-etal-2019-compositional}) as it was shown to be essential to improve compositional generalization.}
    \item \trmb: Meshed-Memory Transformer \cite{cornia2020m2}, a recently proposed Transformer-based architecture that achieves state-of-the-art performance in image captioning on the COCO dataset.
\end{itemize}

\paragraph{Implementation details}
We follow \citet{nikolaus-etal-2019-compositional} and \citet{cornia2020m2} to train their systems.
Model selection is performed using early stopping, which is determined when the \bleu score of the generated captions in the validation set does not increase for five consecutive epochs.\footnote{Whenever present, syntactic tags are stripped away when computing evaluation metrics such as \bleu scores.}
We use the default hyperparameters and do not fine-tune them when tasking the models with syntax generation.
For full experimental details, refer to \cref{sec:app-training}.
Our code and data are publicly available.\footnote{\url{https://github.com/e-bug/syncap}.}
\section{Syntax Awareness}\label{sec:butd}
\begin{table}[t]
    \centering
	\small
	\setlength\tabcolsep{5pt}
    \begin{tabular}{c|lrccccc}
        \toprule
        \multicolumn{2}{l}{\textbf{Model}}     & \textbf{R@5} & \textbf{M} & \textbf{S} & \textbf{C} & \textbf{B} & \textbf{BS} \\
        \midrule
        \multicolumn{2}{l}{\butd} & 9.5 & 25.2 & 18.6 & 92.7 & 32.3 & 41.7 \\
        \midrule
        \multirow{5}{*}{\rotatebox[origin=c]{90}{\plan}}
        & ~+$\idle$ & 8.7 & 23.7 & 17.8 & 87.6 & 30.0 & 38.8 \\
        & ~+$\chunk$ & 10.9 & 24.7 & 18.2 & 89.2 & 31.2 & 41.2 \\
        & ~+$\pos$ & 9.5 & 24.1 & 17.5 & 86.1 & 30.1 & 40.7 \\
        & ~+$\dep$ & 11.1 & 24.6 & 17.8 & 89.7 & 30.8 & 41.0 \\
        & ~+$\ccg$ & 10.6 & 24.5 & 18.0 & 88.4 & 30.4 & 41.0 \\
        \midrule
        \multirow{5}{*}{\rotatebox[origin=c]{90}{\inter}}
        & ~+$\idle$ & 10.5 & 25.3 & 18.8 & 94.3 & 32.3 & 41.7 \\
        & ~+$\chunk$ & 9.7 & 25.2 & 18.7 & 93.4 & 32.5 & 41.7 \\
        & ~+$\pos$ & \textbf{11.8} & 25.4 & 18.8 & 94.4 & \textbf{32.7} & 41.7 \\
        & ~+$\dep$ & 10.8 & 25.2 & 18.7 & 93.0 & 31.9 & 41.6 \\
        & ~+$\ccg$ & 10.5 & 25.4 & \textbf{19.0} & 94.6 & \textbf{32.7} & 41.9 \\
        \midrule
        \multirow{5}{*}{\rotatebox[origin=c]{90}{\multi}}
        & ~+$\idle$ & 9.8 & 25.5 & 18.7 & 94.5 & \textbf{32.7} & 41.8 \\
        & ~+$\chunk$ & 10.3 & 25.5 & \textbf{19.0} & 94.5 & 32.4 & 41.9 \\
        & ~+$\pos$ & 10.3 & 25.4 & 18.8 & 93.8 & 32.6 & 41.8 \\
        & ~+$\dep$ & 11.4 & 25.5 & 18.9 & 93.9 & \textbf{32.7} & 41.9 \\
        & ~+$\ccg$ & 10.8 & \textbf{25.7} & \textbf{19.0} & \textbf{95.6} & \textbf{32.7} & \textbf{42.0} \\
        \bottomrule
    \end{tabular}
    \caption{\label{tab:ablation-syntax} Average validation results for our approaches to integrating syntactic planning into image captioning evaluated across four types of syntactic forms.}
\end{table}

In~\cref{tab:ablation-syntax}, we first report the performance of \butd when jointly modeling different types of syntactic tags and each approach to sentence planning.

\paragraph{Syntax helps compositional image captioning}
\cref{tab:ablation-syntax} clearly shows that, regardless of the level of granularity, syntactic planning enhances compositional generalization in image captioning (R@5).
Moreover, \chunk{} -- one of the most widely-used tag sets for syntax-aware image captioning (\textit{e.g.} \citealt{kuznetsova-etal-2012-collective,9035503}) -- is outperformed by tag sets with finer granularity (e.g. \dep) in \emph{every} approach, motivating further research into incorporating them in image captioning.

Looking at the results for the \plan approach, we see that, with the exception of \pos tags, syntactic planning increases the ability of the model to recall novel concept pairs, with gains of at least $+1.1$ R@5 points.
We then hypothesize that syntax-based sequential planning is effective if the tags convey information about words in relation to each other, e.g. \ccg tags as opposed to \pos tags.

When the model \inter{}s syntactic tags and words, there is an improvement of at least $+1.0$ R@5, except for \chunk. Moreover, \pos tags lead to the highest gain of $+2.3$ R@5.

Finally, the \multi approach also leads to significant gains in compositional generalization, with \dep (original setup of \citealt{currey-heafield-2019-incorporating}) giving the highest R@5, corroborating the effectiveness of our porting into image captioning.

\begin{figure*}[t]
 \centering
 \includegraphics[width=\linewidth, trim={0cm 0cm 0cm 0cm}, clip]{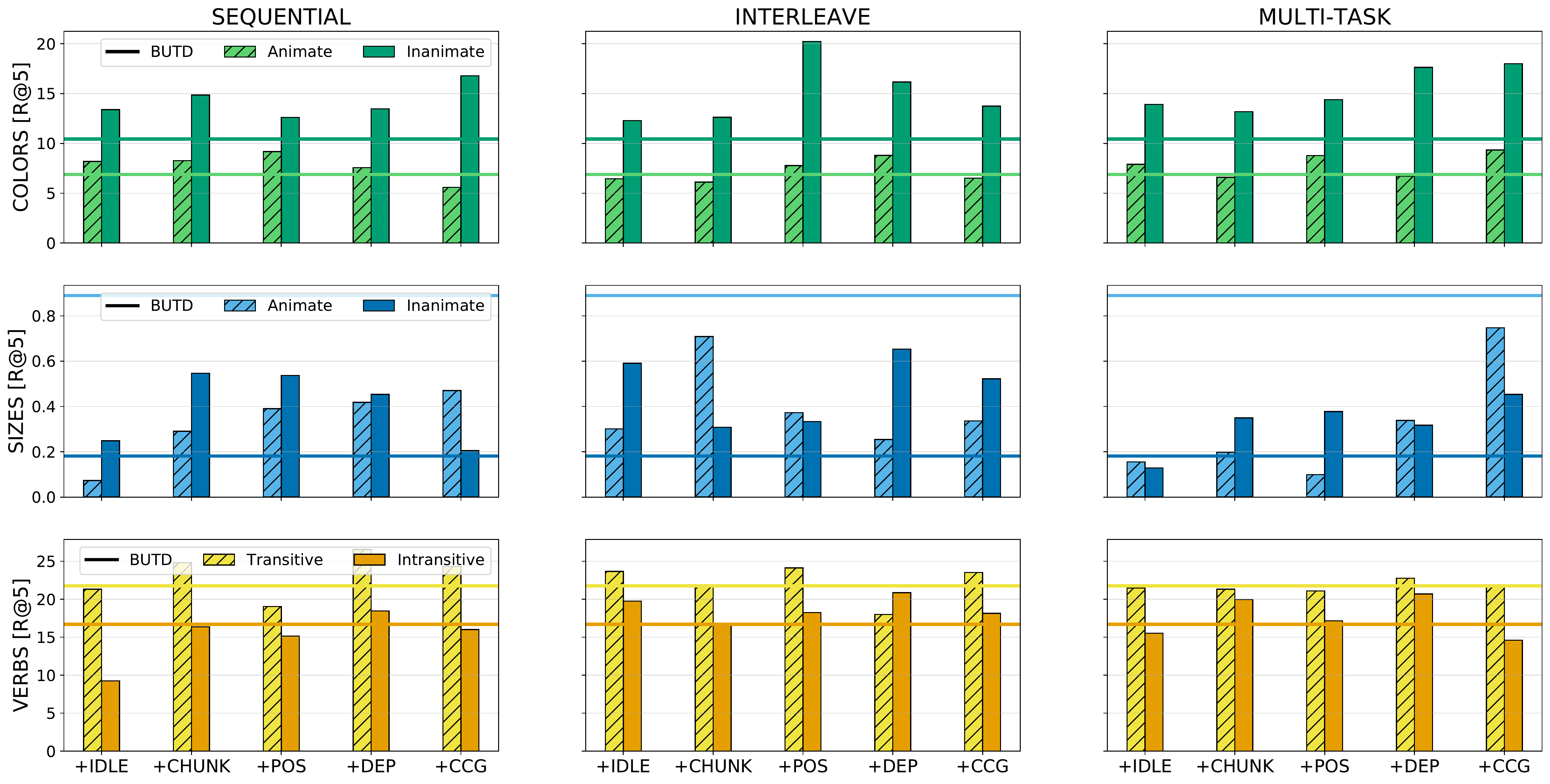}
 \caption{\label{fig:detailed-recall} \rfive of unseen compositions by planning approaches (columns) and composition categories (rows).}
\end{figure*}

\paragraph{Generalization across categories}
We further investigate the role of syntactic planning for the different unseen composition categories defined by \citet{nikolaus-etal-2019-compositional}.
\cref{fig:detailed-recall} illustrates how our different combinations of approaches and syntactic tags deal with color and size, type of the objects (animate and inanimate) and type of the verbs (transitive and intransitive).
We see that \dep tags consistently improve upon \butd for color and size concept pairs, regardless of the planning approach, making them a robust tag set for future research.
\inter{}+\pos also leads to gains for all color and size categories, with up to $+10$ R@5 for colors of inanimate objects.
Conversely, all the variants perform worse than the baseline for the sizes of animate objects.
However, this drop is not substantial because \butd already performs poorly.

\paragraph{Towards neural NLG pipelines}
While the \plan approach closely follows the traditional NLG pipeline, it consistently degrades performance in standard metrics for image captioning.
On the other hand, both \inter and \multi lead to higher performance in compositional generalization and other metrics.
In particular, when \butd is trained to predict either words or \ccg tags in the \multi approach, the generated captions achieve the highest average scores, including a substantial gain of $+2.9$ \cider points.
These results indicate that neural models require novel ways of sentence planning; and that effectively doing so \textit{consistently} leads to the same or better performance in every considered metric.

\paragraph{Grounding the need for planning}
Overall, \cref{tab:ablation-syntax} provides empirical support that an explicit planning step improves compositional generalization in image captioning.
In fact, even breaking bi-grams with the \idlex tag in the \inter approach improves performance: the standard approach of directly mapping image representations to tokens is sub-optimal because the model learns to generate \emph{n}-grams seen during training.

\renewcommand{\arraystretch}{1.1}
\begin{table*}[t]
    \centering
    \small
	\setlength\tabcolsep{5pt}
    \begin{tabular}{lcccccccccccccc}
        \toprule
        & \multicolumn{6}{c}{\textbf{Captioning Evaluation}} & \phantom{ab} & \multicolumn{3}{c}{\textbf{Text Retrieval}} & \phantom{ab} &  \multicolumn{3}{c}{\textbf{Image Retrieval}} \\
                \cmidrule(lr){2-7} \cmidrule(lr){9-11} \cmidrule(lr){13-15}
        \textbf{Model} & \textbf{R@5} & \textbf{M} & \textbf{S} & \textbf{C} & \textbf{B} & \textbf{BS} && \textbf{R@1} & \textbf{R@5} & \textbf{R@10} && \textbf{R@1} & \textbf{R@5} & \textbf{R@10} \\
        \midrule
        \butr & 15.0 & 26.2 & 19.9 & 88.6 & 28.9 & 41.8 && 21.3	& 45.4	& 57.8  && 14.7	& 35.1	& 47.0\\
        ~~~~+\pos & 12.0 & 25.7 & 19.4 & 85.4 & 27.4 & 41.4 && 17.4	& 39.3	& 50.6  && 12.4	& 31.0	& 42.0\\
        \midrule
        \butrmean+\pos & 14.2 & 25.9 & 19.7 & 87.4 & 28.3 & 42.9 && 23.3	& 47.7	& 60.0 && 17.1	& 38.6	& 50.2 \\
        \midrule
        \butrwt & 14.9 & \textbf{26.4} & \textbf{20.2} & 88.8 & 28.5 & \textbf{43.2} && \textbf{26.0}	& \textbf{51.7}	& \textbf{63.6}  && \textbf{18.6}	& \textbf{40.9}	& \textbf{52.8}\\
        ~~~~+\pos & \textbf{16.4} & \textbf{26.4} & 20.0 & \textbf{89.8} & \textbf{29.1} & 43.1 && 24.5	& 48.6	& 60.6  && 18.0	& 40.2	& 52.0\\
        \bottomrule
    \end{tabular}
    \caption{\label{tab:ablation-butr} Average validation results when interleaving syntactic and lexical forms in \butr and our variants.}
\end{table*}

\paragraph{}
Given its superior performance in the recall of novel compositions of concepts, we adopt \inter{}+\pos throughout the remainder of this paper to jointly model syntactic tags and words. For clarity of exposition, we refer to this approach as \pos.

\subsection{Adaptive Re-Ranking for Syntax}

Recall that the best-performing model for compositional image captioning re-ranks its generated captions given the image (\butr; \citealt{nikolaus-etal-2019-compositional}). 
Here, we study how to combine the benefits of syntactic planning and their re-ranking approach. 

The \butd model, investigated above, is a two-layer LSTM \cite{hochreiter1997long} in which the first LSTM encodes the sequence of words, and the second LSTM integrates visual features through an attention mechanism to generate the output sequence~\citep{anderson2018}.
The state-of-the-art \butr{} model extends this with an image–sentence ranking network that projects images and captions into a joint visual-semantic embedding.
The sentence representation used by the ranking network is a learned projection of the final hidden state of the first LSTM: $\mathbf{s} = \mathbf{W} \mathbf{h}^l_{T}.$

\paragraph{Ranking performance}

\cref{tab:ablation-butr} shows that the image--sentence retrieval performance of \butr decreases when interleaving \pos tags with words.
Given the previous formulation of \butr and its connection to \butd, we conclude that jointly modeling syntactic tags and words leads to decreased performance in the generation and ranking tasks.

\begin{table*}[t]
    \centering
	\small
	\setlength\tabcolsep{5pt}
    \begin{subtable}{.5\linewidth}
        \centering
        \begin{tabular}{lrrrrrr}
            \toprule 
            \textbf{Model} & \multicolumn{1}{c}{\textbf{R@5}} & \multicolumn{1}{c}{\textbf{M}} & \multicolumn{1}{c}{\textbf{S}} & \multicolumn{1}{c}{\textbf{C}} & \multicolumn{1}{c}{\textbf{B}} & \multicolumn{1}{c}{\textbf{BS}} \\
            \midrule
            \butd & 9.5 & 25.2 & 18.6 & 92.7 & 32.3 & 41.7 \\
            ~~~~+\pos & 11.8 & 25.4 & 18.8 & 94.4 & 32.7 & 41.7 \\
            \midrule
            \butrwt & 14.9 & 26.4 & 20.2 & 88.8 & 28.5 & 43.2 \\
            ~~~~+\pos & \textbf{16.4} & 26.4 & 20.0 & 89.8 & 29.1 & 43.1 \\
            \midrule
            \trm & 10.6 & 27.9 & 21.6 & \textbf{114.0} & \textbf{37.2} & 44.4 \\
            ~~~~+\pos & 13.2 & \textbf{28.0} & \textbf{21.7} & 113.8 & 35.4 & \textbf{44.9} \\
            \bottomrule
        \end{tabular}
    \end{subtable}%
    \begin{subtable}{.5\linewidth}
        \centering
        \begin{tabular}{lrrrrrr}
            \toprule 
            \textbf{Model} & \multicolumn{1}{c}{\textbf{R@5}} & \multicolumn{1}{c}{\textbf{M}} & \multicolumn{1}{c}{\textbf{S}} & \multicolumn{1}{c}{\textbf{C}} & \multicolumn{1}{c}{\textbf{B}} & \multicolumn{1}{c}{\textbf{BS}} \\
            \midrule
            \butd & 9.2 & 25.4 & 18.6 & 94.4 & 32.4 & 41.8 \\
            ~~~~+\pos & 11.1 & 25.4 & 18.7 & 96.3 & 32.9 & 41.8 \\
            \midrule
            \butrwt & 13.5 & 26.4 & 20.1 & 91.0 & 28.6 & 43.3 \\
            ~~~~+\pos & \textbf{15.4} & 26.3 & 20.0 & 91.0 & 28.7 & 43.2 \\
            \midrule
            \trm & 10.1 & 27.8 & 21.5 & \textbf{115.7} & \textbf{36.5} & 44.5 \\
            ~~~~+\pos & 12.1 & \textbf{28.0} & \textbf{21.6} & \textbf{115.7} & 35.0 & \textbf{44.9} \\
            \bottomrule
        \end{tabular}
    \end{subtable}
    
    \caption{\label{tab:test-results} Average validation (left) and test (right) results when interleaving syntactic (\pos) and lexical forms.}
\end{table*}

\paragraph{Adaptively attending to tags}
We explore two approaches to combining the improvements of interleaved syntactic tagging with ranking:
\begin{itemize}[noitemsep,topsep=1pt]
    \item \textit{mean}: The model creates a mean representation over the hidden states of the first LSTM.
    \item \textit{weight}: The model forms a weighted pooling of the hidden states of the first LSTM layer, whose weights are learned through a linear layer. This is a simple form of attention mechanism and it is equivalent to the one used by~\newcite{nikolaus-etal-2019-compositional} to represent image features in the shared embedding space:
    \begin{align}
    \begin{split}
        {\omega}_t &= \mathbf{W}_\alpha \mathbf{h}^l_{t}, \\
        \boldsymbol{\alpha} &= \text{softmax}\left({\omega} \right), \\
        \mathbf{s} &= \mathbf{W} \sum_{t=1}^{T}{\alpha}_t \mathbf{h}^l_{t}.
    \end{split}
    \end{align}
\end{itemize}

\cref{tab:ablation-butr} shows that the weighting mechanism in the ranking model effectively disentangles syntactic tags and tokens, resulting in $+1.5$ \recall points over \butr, with small improvements to the other metrics.
Compared to \butr, \butrwt also improves the retrieval performance of the ranking module.
Adding \pos tags still decreases retrieval performance but, compared to \butr{}+\pos, the difference is now halved for text retrieval and only $0.7$ points for image retrieval.
Overall, our \butrwt is a more general and robust approach to jointly training a captioning system and a discriminative image--sentence ranker.
\section{Results and Discussion}

We now report the final performance of three image captioning models that integrate the syntactic planning (\inter{}+\pos) with word generation.

\paragraph{Model-agnostic improvements}
We start by investigating whether the compositionality given by syntactic planning generalizes across architectures.
\cref{tab:test-results} reports average validation and test scores for the \butd, \butrwt and \trm models. We find that interleaving \pos tags and tokens consistently leads to $+2$ \recall points in each model without affecting the performance on other metrics, with the exception of decreased \bleu score of \trm.
In this case, \trm+\pos generates captions that are abnormally truncated, ending with bi-grams such as ``of a,'' ``on a'' and ``to a''.\footnote{This is known as reward hacking~\cite{Amodei2016ConcretePI} which arises in models with a reinforcement learning-based optimization phase. Investigating whether proposed approaches to mitigate this problem~\cite[\textit{inter alia}]{liu2017improved,Li_Chen_Liu_2019} are also effective in our setup is left as future work.}
Furthermore, we can clearly see that despite \trm outperforming the RNN-based models in every standard metric, it is only +1 \recall point better than \butd at compositional generalization.
Hence, syntactic planning is an effective strategy to compositional generalization, \emph{regardless} of the language model used.

\paragraph{Generalization across categories}

\begin{table}[t]
    \centering
    \small
    \setlength\tabcolsep{5pt}
    \begin{tabular}{lcccccc}
        \toprule
        \multirow{2}{*}{\textbf{Model}} & \multicolumn{2}{c}{\textbf{Color}} & \multicolumn{2}{c}{\textbf{Size}} & \multicolumn{2}{c}{\textbf{Verb}} \\
        & \multicolumn{1}{c}{\textbf{A}} & \multicolumn{1}{c}{\textbf{I}} & \multicolumn{1}{c}{\textbf{A}} & \multicolumn{1}{c}{\textbf{I}} & \multicolumn{1}{c}{\textbf{T}} & \multicolumn{1}{c}{\textbf{I}} \\
        \midrule
        \butd & 6.9 & 10.4 & \textbf{0.9} & 0.2 & 21.8 & 16.7 \\
        ~~+\pos & 7.8 & 20.2 & 0.4 & 0.3 & 24.1 & 18.2 \\
        \hline
        \butrwt & 15.0 & 24.9 & 0.8 & 1.6 & 26.2 & 20.8 \\
        ~~+\pos & \textbf{16.2} & \textbf{30.4} & \textbf{0.9} & \textbf{2.5} & 26.8 & \textbf{21.9} \\
        \hline
        \trm & 5.2 & 11.9 & 0.2 & 0.2 & 29.1 & 16.6 \\
        ~~+\pos & 7.9 & 17.9 & 0.1 & 0.2 & \textbf{32.3} & 20.6 \\
        \bottomrule
    \end{tabular}
    \caption{Average validation R@5 scores for different categories of held out pairs. Color and size adjectives are split into \textbf{A}nimate or \textbf{I}nanimate objects; Verbs are split into \textbf{T}ransitive and \textbf{I}ntransitive verbs.}
    \label{tab:detailed-recall-val}
\end{table}

\cref{tab:detailed-recall-val} lists the R@5 scores for different categories of held out pairs.
Differently from the results reported by \citet{nikolaus-etal-2019-compositional}, we find the performance of \butd for noun--verb concept pairs to be much higher thanks to a larger beam size (equal to the one used for \butr in our experiments).
Moreover, the performance from interleaving \pos across different categories of held out pairs shows that improvements are consistent across categories and models, with the exception of size modifiers of animate objects, where all models perform poorly.
This was also found by \citet{nikolaus-etal-2019-compositional} and it is likely due to the need for real-world knowledge (i.e. does this image depict a ``big dog'' compared to \emph{all other} ``dogs''?).
For a full breakdown of the R@5 generalization performance for each held out pair by each model, see \cref{tab:recall-pair-val} in \cref{sec:app-analysis}.

\begin{figure}[t]
  \centering
  \includegraphics[width=\linewidth, trim={0cm 0cm 0cm 0cm}, clip]{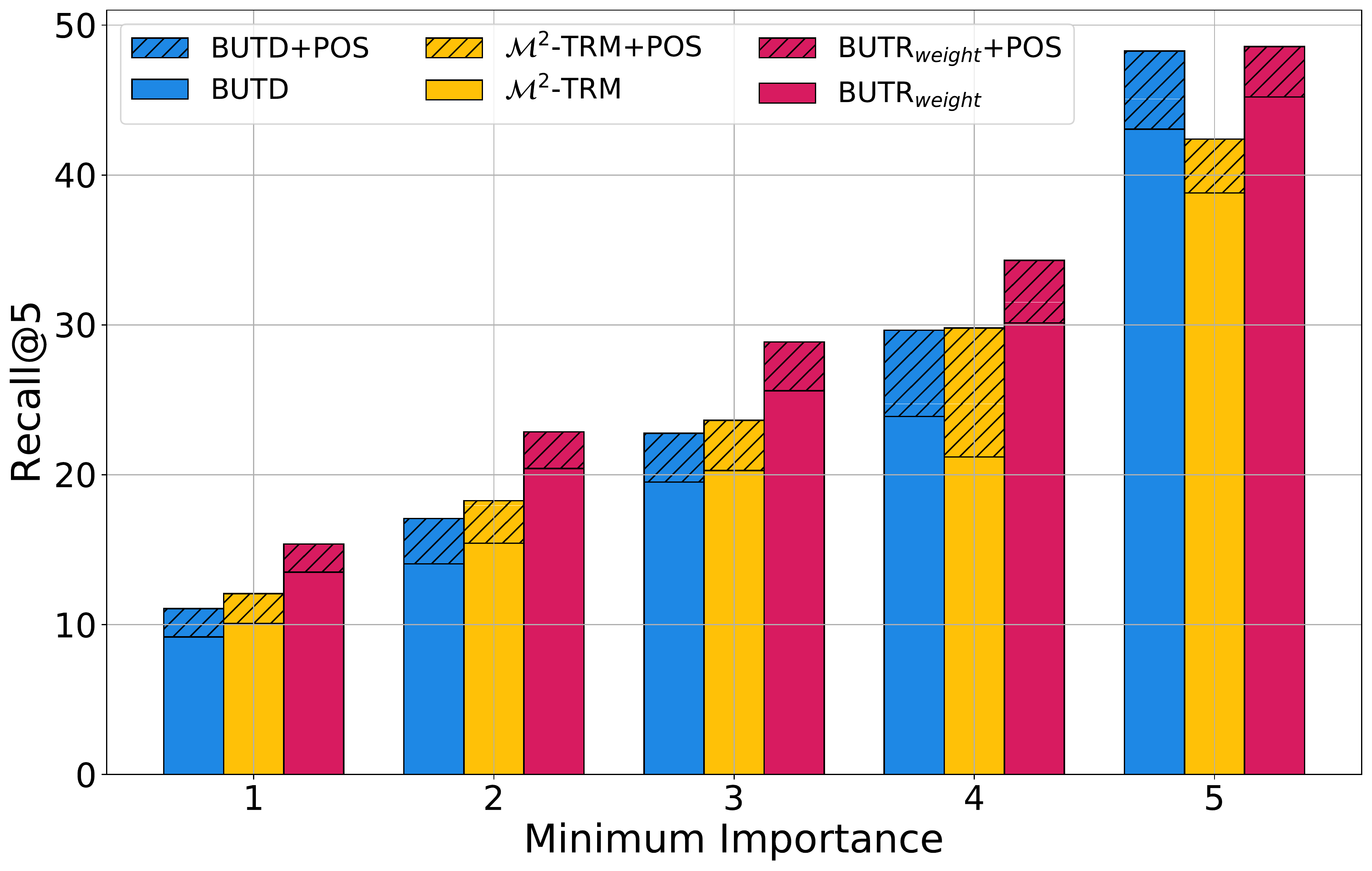}
  \caption{Average test R@5 as a function of the minimum importance of a concept pair for each model.}
  \label{fig:min_imp}
\end{figure}

\paragraph{Performance by minimum importance}
Given that annotators of the COCO dataset were given a relatively open task to describe images, captioning systems should exhibit higher recall of concept pairs when more annotators use them in the descriptions.
As shown in \cref{fig:min_imp}, this behavior is seen in each model, with increasing gains given by jointly modeling lexical and syntactic forms.
In particular, we observe that the \trm model recalls fewer pairs than \butd when they are considered more relevant (more annotators use them in describing an image), and that interleaving \pos tags partially solves its limitations.
Moreover, as agreement among annotators increases, we also see that \butd{}+\pos is as effective as \butrwt, corroborating the effectiveness of our model-agnostic approach against a more complex, multi-task model.

\paragraph{Captions diversity}

\cref{tab:captions-diversity} reports the average scores for caption diversity \cite{van-miltenburg-etal-2018-measuring} in the validation data.
Comparing \butd (RNN-based) and \trm (Transformer-based) models, we see that the output vocabulary of the \trm-based model spans many more word types, resulting in $+11\%$ novel captions.
However, \trm has lower mean segmented type-token ratios (TTRs), contrasting the conclusion of \newcite{van-miltenburg-etal-2018-measuring} that the number of novel descriptions is strongly correlated with the TTR (while this correlation is maintained with the number word types).
The models that jointly model syntactic tags and tokens lead to a higher number of types in both models and a substantial $+8\%$ in novel captions for \trm, without affecting other metrics.
Clearly, \butrwt leads to longer sentences, more types, higher TTRs and the highest percentage of novel captions.
We can also see that \butrwt achieves the highest coverage (defined as the percentage of learnable words it can recall), while \trm has the highest local recall score, being able to better recall the content words that are important to describe a given image.

\begin{table}[t]
    \centering
	\small
	\setlength\tabcolsep{1.7pt}
    \begin{tabular}{lrrrrrrr}
        \toprule 
        \textbf{Model} &   \textbf{ASL}  &  \textbf{Types}  &  \textbf{TTR}$_1$  &  \textbf{TTR}$_2$  &  \textbf{$\%$Novel}  &  \textbf{Cov}  &  \textbf{Loc}$_5$ \\
        \midrule
        \butd  &  8.6  &  463  &  0.16  &  0.37  &  69.2  &  0.12  &  0.74 \\
        ~~~~+\pos  &  8.6  &  466  &  0.16  &  0.37  &  70.3  &  0.12  &  0.75 \\
        \butrwt  &  \textbf{10.3}  &  \textbf{783}  &  \textbf{0.20}  &  \textbf{0.49}  &  \textbf{97.2}  &  \textbf{0.20}  &  0.78 \\
        ~~~~+\pos  &  \textbf{10.3}  &  778  &  \textbf{0.20}  &  \textbf{0.49}  &  96.7  &  \textbf{0.20}  &  0.78 \\
        \trm  &  9.1  &  580  &  0.14  &  0.33  &  80.1  &  0.15  &  \textbf{0.83} \\
        ~~~~+\pos  &  9.5  &  601  &  0.13  &  0.33  &  88.4  &  0.15  &  \textbf{0.83} \\
        \bottomrule
    \end{tabular}
    \caption{\label{tab:captions-diversity} Average validation set scores for diversity metrics as defined in \newcite{van-miltenburg-etal-2018-measuring}.}
\end{table}

\begin{table*}[t]
    \centering
    \small
    \begin{tabular}{m{3.5cm} m{11.5cm}}
            \includegraphics[width=0.14\textwidth, margin=.5em, clip]{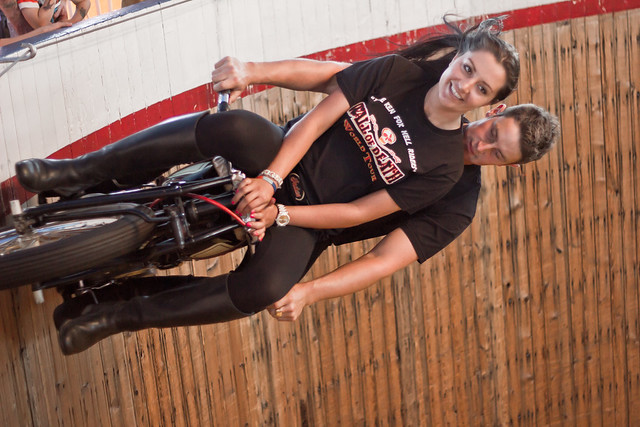}
            &
            \begin{itemize}[noitemsep,topsep=1pt,leftmargin=-1pt]
                \item[]
                \item[] \butd: there is a woman that is on the floor
                \item[] \butd+ \plan: a woman doing a trick on a bicycle
                \item[] \butd+ \inter: a \underline{woman riding} a bike on a wooden floor
                \item[] \butd+ \multi: a \underline{woman riding} a bike on a wooden surface
            \end{itemize}
            \\[25pt]
            \includegraphics[width=0.14\textwidth, margin=.5em, trim={0cm 4.5cm 0cm 1cm}, clip]{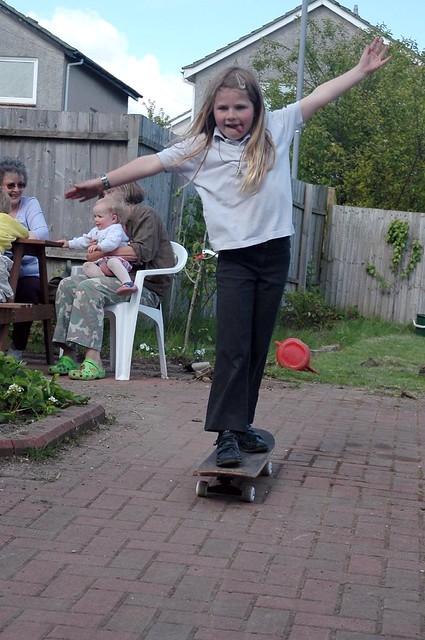}
                &
                \begin{itemize}[noitemsep,topsep=1pt,leftmargin=-1pt]
                \item[]
                \item[] \butd: a woman with a child sitting on a bench
                \item[] \butd+ \pos : \textbf{a girl that is standing}  on a skateboard

                \vspace{8pt}
                \item[] \butrwt : a girl and child playing with a toy in a backyard
                \item[] \butrwt+ \pos : \textbf{a girl doing a trick on a skateboard} on a brick walkway
                
                \vspace{8pt}
                \item[] \trm : a girl doing a trick on a skateboard with
                \item[] \trm+ \pos : a little girl \textbf{standing on a skateboard} in the
                \vspace{-4pt}
                \end{itemize}
                \\[-6pt]
    \end{tabular}
        \captionof{figure}{\textbf{Top:} Captions generated by \butd with different approaches to integrating syntactic tags (based on best \rfive). \textbf{Bottom:} Captions generated by \butd, \butrwt, and \trm, and each model when interleaving \pos tags. Syntax-aware approaches generate \textbf{higher quality} captions with more \underline{unseen concept pairs}.
        }
    \label{tab:qual-examples}
    \vspace{-8pt}
\end{table*}

\paragraph{Accuracy of syntactic forms}
We verify that the models can correctly predict syntactic tags, regardless of their granularity and the approach used to jointly modeling them with tokens.
Indeed, the accuracy of the generated syntactic tags, measured as the ratio of \textit{sequences} matching the annotations from StanfordNLP, by \butd is high, ranging between $95\%$ and $99\%$.
See \cref{sec:app-analysis} for details.

\paragraph{Qualitative examples}
\cref{tab:qual-examples} shows generated captions.
Compared to standard \butd, all syntax-aware approaches allow the model to recall more \underline{unseen concept pairs}, while also improving the overall quality of the captions.
In addition, when looking at the captions generated by all three models, both with and without interleaving \pos tags, we find the integration of syntactic tags to clearly \textbf{improve the quality} of the generated caption. 
See \cref{tab:more-qual-examples} in \cref{sec:app-analysis} for more examples.
\section{Related Work}
\paragraph{Compositional image captioning}
\newcite{nikolaus-etal-2019-compositional} studies compositional generalization in image captioning with combinations of unseen adjective--noun and verb--noun pairs, whose constituents are observed at training time but not their combination, thus introducing a \textit{paradigmatic gap} in the training data.
\newcite{nikolaus-etal-2019-compositional} showed how to improve compositional generalization by jointly training an image--sentence ranking model with a captioning model. Other work has also investigated generalization to unseen combinations of visual concepts as a classification task \cite{misra2017red,kato2018compositional}, triplet prediction \cite{Atzmon2016LearningTG}, or unseen objects \cite{lu2018neural}. 
Here, we improve generalization by jointly modeling syntactic tags and tokens, and we show how to combine this with the improvements gained from a jointly-trained ranking model.

\paragraph{Joint syntactic and semantic representations}
While little work has investigated the interaction of jointly modeling semantics and various syntactic forms in captioning models, a few studies have exploited syntax in image and video captioning.
\newcite{ijcai2018-0168} propose a multi-task system to jointly train the task of image captioning with two additional tasks: multi-object classification and syntax generation. 
The same LSTM decoder is used to generate captions \textit{and} \ccg tags by mapping the hidden representations to either word or tag vocabularies through two different output layers.
\newcite{dai2018neural} propose a two-stage sequential pipeline where a sequence of noun-phrases is first selected from a fixed pool, which are then patched together via predetermined connecting phrases. 
This method, however, is unlikely to realize any benefits for compositional generalization because it uses the top-$50$ noun-phrases and $1,000$ connecting phrases from the training set.
Our \inter approach can be used to address these limitations in their ``phrase pool'' and ``connecting'' modules to produce unseen compositions.
\newcite{deshpande2019fast} rely on sequences of \pos tags to produce diverse captions.
Similarly to our \plan approach, their model first predicts a sequence of \pos tags conditioned on the input image. 
However, the authors limit the \pos sequences to $1,024$ templates obtained through quantization of the training set.
During inference, the model samples $k$ \pos tag sequences and uses them to condition a greedy decoder for captions generation.
\newcite{hou2019joint} take yet another approach to jointly learn \pos tags and surface forms in the framework of video captioning.
They introduce a model that resembles our \inter approach but with two main differences: (i) the $t$-th tag is not conditioned on previous tags, and (ii) the $t$-th word is only conditioned on the $t$-th tag and the video.
\section{Conclusion}

We investigated a variety of approaches along with the use of syntactic tag sets to achieve compositional generalization in image captioning via sentence planning.
Our results support the claim that combining syntactic planning and language generation \emph{consistently} improves the generalization capability of RNN- and Transformer-based image captioning models, especially for inanimate color--noun combinations.
While this approach penalizes image--sentence ranking models, we showed that this can be overcome with an adaptive mechanism, resulting in state-of-the-art performance on the compositional generalization task.
We believe our results will lead to further exploration of syntax-aware captioning models given their potential to better generalize, both in terms of under-researched syntactic granularity (e.g. \ccg) and more expressive alternatives to modeling syntactic structure. 
Another direction for future work is to focus on size--noun compositions, which rely on the successful integration of real-world knowledge.

\section*{Acknowledgments}
{\scriptsize\euflag}
We are grateful to the anonymous reviewers, Mitja Nikolaus, Mert Kiliçkaya and members of the CoAStaL NLP group for their useful comments and discussions.
This project has received funding from the European Union's Horizon 2020 research and innovation programme under the Marie Sk\l{}odowska-Curie grant agreement No 801199.

\bibliography{eacl2021}
\bibliographystyle{acl_natbib}

\clearpage
\appendix

\section{Experimental Setup} \label{sec:app-training}

\begin{table*}[t]
    \centering
    \begin{tabular}{llrrr}
        \toprule 
        & \textbf{Held out pairs} & $\boldsymbol{\mathcal{D}_{train}}$ & $\boldsymbol{\mathcal{D}_{val}}$ & $\boldsymbol{\mathcal{D}_{test}}$ \\
        \midrule
        1 & black cat, big bird, red bus, small plane, eat man, lie woman & $79,847$ & $2,936$ & $1,349$ \\
        2 & brown dog, small cat, white truck, big plane, ride woman, fly bird & $79,823$ & $2,960$ & $1,355$ \\
        3 & white horse, big cat, blue bus, small table, hold child, stand bird & $79,922$ & $2,861$ & $1,459$ \\
        4 & black bird, small dog, white boat, big truck, eat horse, stand child & $79,627$ & $3,156$ & $1,501$ \\
        \bottomrule
    \end{tabular}
    \caption{\label{tab:datasets} Held out word pairs in each dataset split. Dataset sizes are measured in number of images.}
\end{table*} 

\begin{table*}[t]
    \centering
    \small
    \begin{minipage}{0.5\linewidth}
    \centering
    \begin{tabular}{lccc}
        \toprule
        \textbf{Approach} &  \textbf{T} & d(S$_t$, W$_t$) & \textbf{N} \\
         \midrule
        {\plan} & 2$\times$ & T & 1$\times$ \\
        {\inter} & 2$\times$ & 1 & 1$\times$ \\
        {\multi} & 1$\times$ & -- & 2$\times$ \\
        \bottomrule
    \end{tabular}
    \caption{Comparison of the planning approaches studied in this paper in terms of the sequence lengths used to train the captioning models (\textbf{T}), the distance between the syntactic tag and its corresponding word d(S$_t$, W$_t$), and the number of training examples per epoch (\textbf{N}).}
    \label{tab:plan_properties}
    \end{minipage}%
    \begin{minipage}{0.5\linewidth}
    \centering
    \begin{tabular}{lr}
    \toprule
        \textbf{Model} & \textbf{Training time [hour]} \\ 
    \midrule
        \butd & 6 \\
        ~~~~+\pos & 9 \\
        \butrwt & 18 \\
        ~~~~+\pos & 17 \\
        \trm & 383\\
        ~~~~+\pos & 414\\
    \bottomrule
    \end{tabular}
    \caption{Training times on held out dataset 2.}
    \label{tab:model_details}
    \end{minipage}
\end{table*}

\begin{table*}[t]
    \centering
    \begin{tabular}{l|cc|cc|cc}
        \toprule
         \textbf{Concept pair} & \textbf{\butd} & \textbf{+\pos} & \textbf{\butrwt} & \textbf{+\pos} & \textbf{\trm} & \textbf{+\pos} \\
        \midrule
        black cat & 16.1 & 15.3 & 26.0 & 29.1 & 5.0 & 10.9 \\
        eat man & 19.0 & 24.5 & 26.5 & 26.5 & 23.8 & 29.1 \\
        small plane & 0.2 & 0.0 & 0.0 & 0.4 & 0.6 & 0.0 \\
        red bus & 16.1 & 24.0 & 51.8 & 54.2 & 13.3 & 24.9 \\
        big bird & 2.4 & 0.9 & 0.9 & 0.9 & 0.5 & 0.0 \\
        lie woman & 17.5 & 18.5 & 27.1 & 20.8 & 14.9 & 19.8 \\
        ride woman & 27.3 & 30.2 & 23.5 & 27.4 & 28.5 & 30.8 \\
        white truck & 17.6 & 30.6 & 23.3 & 29.4 & 9.4 & 12.7 \\
        fly bird & 21.8 & 25.5 & 26.3 & 32.5 & 29.2 & 35.8 \\
        small cat & 1.2 & 0.4 & 1.2 & 1.6 & 0.4 & 0.0 \\
        brown dog & 1.7 & 0.5 & 4.8 & 7.8 & 0.9 & 1.2 \\
        big plane & 0.5 & 0.6 & 3.7 & 8.7 & 0.3 & 0.0 \\
        stand bird & 18.4 & 18.6 & 16.9 & 24.0 & 16.8 & 19.5 \\
        white horse & 4.9 & 10.9 & 16.9 & 21.3 & 10.1 & 10.1 \\
        small table & 0.0 & 0.0 & 0.0 & 0.0 & 0.0 & 0.0 \\
        big cat & 0.0 & 0.0 & 0.0 & 0.0 & 0.0 & 0.0 \\
        blue bus & 6.5 & 22.7 & 21.3 & 33.2 & 23.5 & 32.1 \\
        hold child & 17.8 & 19.8 & 10.3 & 13.8 & 14.8 & 13.0 \\
        big truck & 0.0 & 0.7 & 2.9 & 0.7 & 0.0 & 1.0 \\
        white boat & 1.6 & 3.5 & 3.3 & 4.9 & 1.6 & 1.9 \\
        small dog & 0.0 & 0.1 & 1.2 & 0.9 & 0.0 & 0.3 \\
        eat horse & 23.0 & 22.1 & 44.6 & 39.4 & 49.3 & 56.3 \\
        stand child & 9.1 & 10.4 & 13.0 & 10.3 & 5.7 & 7.4 \\
        black bird & 4.9 & 4.4 & 12.3 & 6.4 & 4.9 & 9.3 \\
        \bottomrule
    \end{tabular}
    \caption{R@5 for each of the held out concept pairs in the validation sets.}
    \label{tab:recall-pair-val}
\end{table*}

\begin{table*}
    \begin{tabular}{m{3.5cm} m{11cm}}
            \includegraphics[width=0.14\textwidth, margin=.5em, trim={0cm 4cm 0cm 2cm}, clip]{figures/eat-man.jpg}
            &
            \begin{itemize}[noitemsep,topsep=1pt,leftmargin=-1pt]
                \item[] \butd: a man standing in front of a pizza
                \item[] \butd+ \plan: a couple of people that are standing around a pizza
                \item[] \butd+ \inter: a man \emph{sitting} at a table with a pizza
                \item[] \butd+ \multi: a man \emph{sitting} at a table with a pizza
            \end{itemize}
            \\
        \midrule
            \includegraphics[width=0.14\textwidth, margin=.5em]{figures/lie-woman.jpg}
            &
            \begin{itemize}[noitemsep,topsep=1pt,leftmargin=-1pt]
                \item[]
                \item[] \butd: a baby laying on a bed with a teddy bear
                \item[] \butd+ \plan: a couple of kids sitting on a bed
                \item[] \butd+ \inter: a large teddy bear sitting on top of a bed
                \item[] \butd+ \multi: two children sitting on a bed with a laptop
            \end{itemize}
            \\
        \midrule
            \includegraphics[width=0.14\textwidth, margin=.5em]{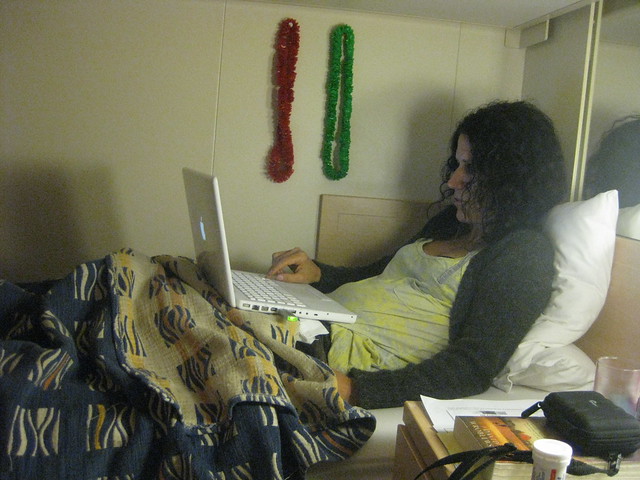}
            &
            \begin{itemize}[noitemsep,topsep=1pt,leftmargin=-1pt]
                \item[]
                \item[] \butd: a man sitting on a bed using a laptop computer
                \item[] \butd+ \pos : a person sitting on a bed with a laptop
                
                \vspace{8pt}
                \item[] \butrwt: a woman sitting on a bed and using a laptop
                \item[] \butrwt+ \pos : a woman sitting on a bed with a laptop and a laptop
                
                \vspace{8pt}
                \item[] \trm: a man sitting on a couch using a laptop computer
                \item[] \trm+ \pos : a woman sitting on a bed using a laptop computer
                \vspace{-5pt}
            \end{itemize}
            \\
        \midrule
            \includegraphics[width=0.14\textwidth, margin=.5em, trim={10cm 0cm 0cm 0cm}, clip]{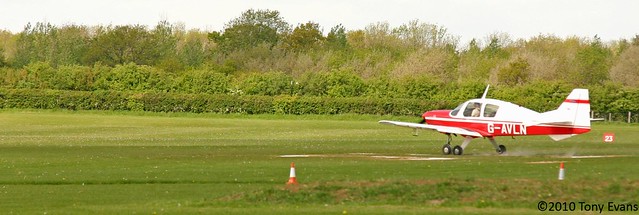}
            &
            \begin{itemize}[noitemsep,topsep=1pt,leftmargin=-1pt]
                \item[]
                \item[] \butd: a plane sitting on top of a lush green field
                \item[] \butd+ \pos : a red and white plane in an open field
                
                \vspace{8pt}
                \item[] \butrwt: a red and white plane taking off on a field
                \item[] \butrwt+ \pos : a red and white plane on a lush green field
                
                \vspace{8pt}
                \item[] \trm: an airplane is on the runway in a field
                \item[] \trm+ \pos : a red and white plane sitting on a runway
                \vspace{-5pt}
            \end{itemize}
            \\
        \midrule
            \includegraphics[width=0.14\textwidth, margin=.5em]{figures/puppy-kitty.jpg}
            &
            \begin{itemize}[noitemsep,topsep=1pt,leftmargin=-1pt]
                \item[]
                \item[] \butd: a couple of sheep standing on top of a lush green field
                \item[] \butd+ \pos : a couple of animals standing in the grass
                
                \vspace{8pt}
                \item[] \butrwt: two small lambs standing in a green field
                \item[] \butrwt+ \pos : two baby bears standing in a grassy field
                
                \vspace{8pt}
                \item[] \trm: a mother sheep and a baby sheep in a field
                \item[] \trm+ \pos : a baby sheep standing next to an adult sheep in
                \vspace{-5pt}
            \end{itemize}
            \\
    \end{tabular}
    \captionof{figure}{More examples of generated captions from the validation sets. While syntax-aware approaches generate more accurate captions overall, they are sometimes worse than the standard system (second example). Moreover, the last example shows how most systems confuse a kitty and a puppy with two sheep, lambs or bears.}
    \label{tab:more-qual-examples}
\end{table*}

\paragraph{Data}
In order to evaluate the compositional generalization of a model, we use training and evaluation sets such that paradigmatic gaps are observed in the training set.
That is, for a concept pair $\{c_i, c_j\}$, the validation $\mathcal{D}_{val}$ and test $\mathcal{D}_{test}$ sets only contain images in which at least one of the captions contains the pair of concepts, while the complementary set -- where concepts $c_i$ and $c_j$ can only be observed independently -- is used for training $\mathcal{D}_{train}$.
Specifically, following \citet{nikolaus-etal-2019-compositional}, we select the same $24$ adjective--noun and verb--noun concept pairs, and split the English COCO dataset~\cite{lin2014microsoft} into four sets, each containing six held out concept pairs (training and validation instances are drawn from \texttt{train2014}, while test instances from \texttt{val2014}).
\cref{tab:datasets} lists the sizes (in number of images) of each split.\footnote{Note that the size of each set is slightly different from the one in \citet{nikolaus-etal-2019-compositional} as they used different tools for tokenizing and parsing, while we use a single framework to maximize its performance when parsing the captions (used to identify concept pair candidates).}
For more details, we refer the reader to \citet{nikolaus-etal-2019-compositional}.

\paragraph{Training details}

Following \citet{nikolaus-etal-2019-compositional} and \citet{cornia2020m2}, each system is trained with teacher forcing. 
Model selection is performed using early stopping, which is determined when the \bleu score of the generated captions in the validation set does not increase for five consecutive epochs.\footnote{When present, syntactic forms are stripped away when computing evaluation metrics such as \bleu scores.}
All models are trained using the Adam optimizer~\citep{kingma2014adam}: \butd and \butr use an initial learning rate of $1e-4$, $\beta_1=0.9$ and $\beta_2=0.999$, and gradients are clipped when they exceed $10.0$.
For the GradNorm optimizer~\cite{pmlr-v80-chen18a} used in \butr, the initial learning rate is $0.01$ and the asymmetry is $2.5$, although we find it beneficial to tune the latter when generating syntax.\footnote{Searched over the minimal grid: $\alpha \in \{1.5, 2.0, 2.5, 4.0\}$.} 
Moreover, we find that taking the absolute value of the GradNorm weights for each loss in the renormalization step (given that our loss functions are by definition positive) leads to more stable multi-task training.
\trm first uses an initial learning rate of $1$, $\beta_1=0.9$ and $\beta_2=0.98$, with a warm-up equal to $10,000$ iterations \cite{vaswani2017attention}, and it is then fixed to $5e-6$ during \ciderd optimization.
A batch size of $50$ is used when training \trm, while \butd and \butr are trained with batch sizes of $100$.
When adding syntactic forms, due to memory limitations, a batch size of $50$ is always used (see \cref{tab:plan_properties} for a comparison of the planning approaches).
All models are trained on one NVIDIA TitanX GPU in a shared cluster. 
See \cref{tab:model_details} for running times on the second held out dataset.

\paragraph{Inference}
At evaluation time, a maximum caption length of $20$ is used when generating lexical forms only, and of $40$ when also syntactic tags are generated.
Notably, we use the default hyperparameters provided by the respective authors and do not fine-tune them when tasking the models with syntax generation.
Differently from \citet{nikolaus-etal-2019-compositional}, rather than using a beam of $100$ for \butr only, we let all systems generate captions using such beam size as we found it to significantly improve compositional generalization of \butd in our validation sets.\footnote{\butd \rfive: $7.8$ (beam size $5$), $9.5$ (beam size $100$).}

\section{Further Analysis} \label{sec:app-analysis}

\begin{figure}
 \centering
 \includegraphics[width=\linewidth, trim={0cm 0cm 0cm 0cm}, clip]{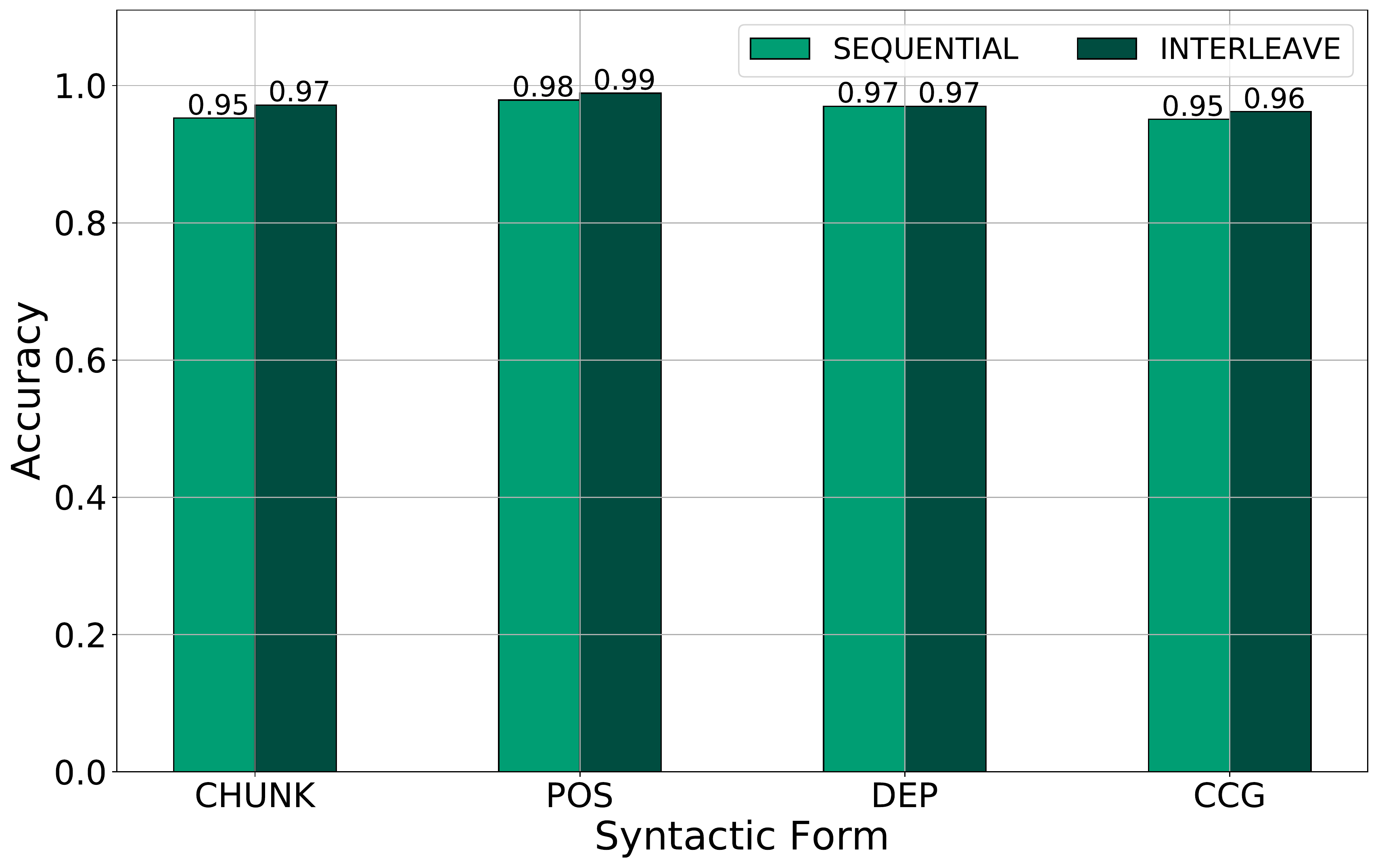}
 \caption{Validation accuracy of syntactic tags generated by \butd using \plan and \inter.}
 \label{fig:tas_acc}
\end{figure}

\paragraph{Accuracy of syntactic forms}
We verify that a model can correctly predict syntactic forms, regardless of their granularity and the approach used to jointly modeling them with lexical forms.
\cref{fig:tas_acc} shows that, indeed, the accuracy of the generated syntactic tags, measured as the ratio of \textit{sequences} matching the annotations from StanfordNLP, by \butd is high, ranging between $95\%$ and $99\%$.
Note that we only evaluate accuracy of the \plan and \inter approaches as there is no close relationship between syntactic and lexical sequences in the \multi approach.

\paragraph{Qualitative examples}
\cref{tab:more-qual-examples} shows more generated captions for images in the validation sets.

\end{document}